\documentclass[letterpaper]{article} 
\usepackage{aaai25}  
\usepackage{times}  
\usepackage{helvet}  
\usepackage{courier}  
\usepackage[hyphens]{url}  
\usepackage{graphicx} 
\usepackage{natbib}  
\usepackage{caption} 
\usepackage{algorithm}
\usepackage{algorithmic}

\usepackage{booktabs}
\usepackage{multirow}
\usepackage{colortbl, booktabs}
\usepackage{amsmath}
\usepackage{xcolor}
\usepackage{amssymb}
\usepackage{subfigure}

\usepackage{newfloat}
\usepackage{listings}
\DeclareCaptionStyle{ruled}{labelfont=normalfont,labelsep=colon,strut=off} 
\floatstyle{ruled}
\newfloat{listing}{tb}{lst}{}
\floatname{listing}{Listing}
\title{RI-MAE: Rotation-Invariant Masked AutoEncoders for Self-Supervised\\ Point Cloud Representation Learning}
\author{
    Kunming Su\textsuperscript{\rm 1}, Qiuxia Wu\textsuperscript{\rm 1}\thanks{Corresponding author}, Panpan Cai\textsuperscript{\rm 1}, Xiaogang Zhu\textsuperscript{\rm 2}, \\Xuequan Lu\textsuperscript{\rm 3}, Zhiyong Wang\textsuperscript{\rm 4}, Kun Hu\textsuperscript{\rm 4}
}
\affiliations{
    \textsuperscript{\rm 1}South China University of Technology,
    \textsuperscript{\rm 2}The University of Adelaide,\\
    \textsuperscript{\rm 3}The University of Western Australia,
    \textsuperscript{\rm 4}The University of Sydney\\
    kunmingsu07@gmail.com, \{qxwu, caipp\}@scut.edu.cn, xiaogang.zhu@adelaide.edu.au, \\bruce.lu@uwa.edu.au, \{zhiyong.wang, kun.hu\}@sydney.edu.au
}

\usepackage{bibentry}

\begin{document}

\maketitle

\begin{abstract}
Masked point modeling methods have recently achieved great success in self-supervised learning for point cloud data. However, these methods are sensitive to rotations and often exhibit sharp performance drops when encountering rotational variations. In this paper, we propose a novel Rotation-Invariant Masked AutoEncoders (RI-MAE) to address two major challenges: 1) achieving rotation-invariant latent representations, and 2) facilitating self-supervised reconstruction in a rotation-invariant manner. For the first challenge, we introduce RI-Transformer, which features disentangled geometry content, rotation-invariant relative orientation and position embedding mechanisms for constructing rotation-invariant  point cloud latent space. For the second challenge, a novel dual-branch student-teacher architecture is devised. It enables the self-supervised learning via the reconstruction of masked patches within the learned rotation-invariant latent space. Each branch is based on an RI-Transformer, and they are connected with an additional RI-Transformer predictor. The teacher encodes all point patches, while the student solely encodes unmasked ones. Finally, the predictor predicts the latent features of the masked patches using the output latent embeddings from the student, supervised by the outputs from the teacher. Extensive experiments demonstrate that our method is robust to rotations, achieving the state-of-the-art performance on various downstream tasks. Our code is available at https://github.com/kunmingsu07/RI-MAE .
\end{abstract}

\section{Introduction}

\begin{figure}
	\centering
			\includegraphics[width=0.95\linewidth]{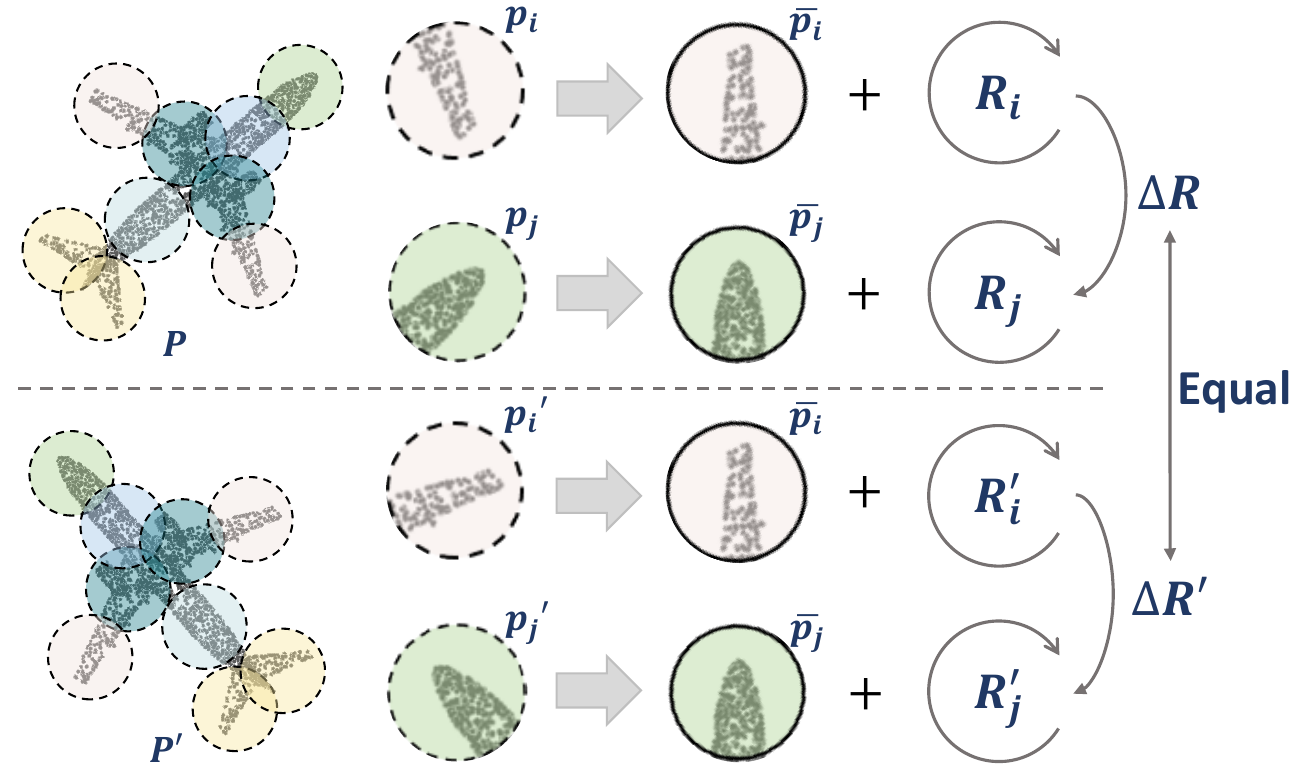}
	\caption{Illustration of the consistent relative orientation between point patches. A point cloud can be partitioned into several point patches, each of which can be regarded as an aligned patch rotated from a canonical pose with a specific rotation. While the overall pose of the point cloud can undergo various rotations, the relative rotation $\Delta R$ between any two patches remains constant and thus is rotation-invariant.}
	\label{fig:fig1}
\end{figure}

Pioneered by PointNet \cite{qi2017pointnet}, deep neural networks have achieved great success in point cloud representation learning, and shown impressive results on various downstream tasks such as point cloud classification and segmentation \cite{qi2017pointnet++,wang2019dynamic,li2018pointcnn}. However, such approaches are supervised and necessitate data annotated with labor-intensive efforts for training models. In contrast, self-supervised learning methods learn to derive representation from unlabeled data by designing pretext tasks, achieving notable  success in the fields of natural language processing (NLP) and computer vision (CV). Inspired by this success, many self-supervised point cloud representation learning approaches have been proposed \cite{yang2018foldingnet,Liu2019L2GAU,afham2022crosspoint,Chen2021ShapeSF,Poursaeed2020SelfSupervisedLO,rao2020global,sanghi2020info3d,Tsai2022SelfSupervisedFL,sun2022self}. They learn to acquire point cloud representations through a variety of tasks, such as diverse forms of point cloud reconstruction and contrastive learning tasks.

Recently, building on the success of masked language modeling (MLM) and masked image modeling (MIM),  as exemplified by BERT \cite{devlin2018bert} and Masked
AutoEncoder (MAE) \cite{He2021MaskedAA}, several studies have introduced transformer-based masked point modeling (MPM) approaches \cite{yu2022point,pang2022masked,zhang2022point,dong2023act,zhang2023learning,qi2024shapellm,chen2024pointgpt,qi2023recon}. They divide the input point cloud into point patches and apply a substantial masking ratio to randomly mask these patches. Subsequently, a transformer-based architecture is utilized to encode the unmasked patches and to reconstruct the masked ones, serving as a self-supervised task in the pretraining phase. Thanks to the capabilities of the transformer model and the strategically designed pretext tasks, the encoder is capable of capturing high-level point cloud features from partially unmasked data. This process leads to remarkable enhancements in performance on downstream tasks through fine-tuning.

Nonetheless, existing MPM methods are sensitive to rotations. They overlook the effects of point cloud orientation, assuming that the input point clouds are pre-aligned. This assumption does not hold in real-world scenarios, where point clouds typically involve arbitrary rotations due to varying capture settings. As a result, the performance on downstream tasks significantly drops when faced with rotated point cloud data. To address this essential problem, this study aims to devise a rotation-invariant MPM approach for point cloud data. We begin by identifying and analyzing two primary challenges to integrate rotation invariance into MPM:

\begin{enumerate}
    \item \textit{How to achieve rotation-invariant latent representations?} 
    Existing methods attain rotation invariance by transforming the input point cloud into handcrafted features \cite{chen2019clusternet,li2021rotation}, designing rotation-invariant convolutions \cite{chen2022devil,zhang2022riconv++}, or establishing global reference frames \cite{zhao2022rotation,li2021closer}. Nevertheless, directly integrating these solutions into transformer-based MPM poses significant challenges, as they demand specialized network designs that are inherently incompatible with the transformer framework. 
    Despite the introduction of a rotation-invariant transformer in \cite{yu2023rethinking}, its complex architecture hinders the development of a universally applicable MPM framework.
    Hence, instead of making extensive modifications to the Transformer, we aim to achieve rotation invariance with minimal modifications, 
    preserving the advantages of transformer-based MPM architectures.
    \item \textit{How to reconstruct masked patches in a rotation-invariant manner?} 
    Conventional self-supervised reconstruction strategy for MPM is directly guided by the the raw input point cloud, which contains orientation patterns and is not applicable for constructing a rotation-invariant MPM approach. 
    For two point clouds that are identical in content but differ in orientation, their learned latent representations should be identical. However, a conventional reconstruction process is expected to yield two distinct results when supervision is applied to revert them back to their original input orientations. 
    Such ambiguous supervision for reconstruction causes discrepancies in losses and less desired reconstructions.
    As such,  our goal is to devise effective reconstruction mechanisms that is not within the raw input space.
\end{enumerate}

Based on these insights, we propose a novel Rotation Invariant Masked AutoEncoders for self-supervised learning on point clouds, namely RI-MAE. 
Addressing the first challenge, we propose a rotation-invariant transformer, RI-Transfomer, for point clouds, as the backbone of RI-MAE. Specifically, 
a content-orientation disentanglement mechanism is introduced to align point patches to their canonical poses for rotation-invariant representations. 
To assist this representation learning process, a rotation-invariant orientation embedding (RI-OE) and a rotation-invariant position embedding (RI-PE) are proposed. 
RI-OE leverages the rotation-invariant nature of the relative orientation between two patches as illustrated in Fig. \ref{fig:fig1}. The orientation of a point patch is defined by a rotation matrix relative to its canonical pose. 
RI-PE adopts the center coordinates and rotation matrix of a point patch to address the issue of conventional rotation-sensitive point cloud position embeddings.  
With these compatible mechanisms, our RI-Transfomer inherently supports rotation invariance, enhancing the usage of general transformers for the learning on point clouds.


For the second challenge, we propose a novel
dual-branch student-teacher architecture for self-supervised reconstruction of masked point patches within a learned rotation-invariant latent space shared by the two branches. 
Specifically, the dual-branch consists of two RI-Transformers, each corresponding to one branch, and they are connected with an additional RI-Transformer predictor. 
The teacher branch processes and encodes all point patches, while the student branch solely addresses
 the visible set.
Subsequently, the predictor is employed to predict the features of the masked point patches using the student latent embeddings of visible point patches, along with the supervision from the teacher branch's outputs. 


Our main contributions are summarized as follows:
\begin{enumerate}

    \item We propose RI-Transfomer with disentangled geometry content, and rotation-invariant patch orientation and position embedding mechanisms to learn rotation-invariant latent representations for point clouds. 
    \item We propose a dual-branch architecture, RI-MAE, based on the RI-Transformer backbone, enabling self-supervised masked patch reconstruction within a learned rotation-invariant latent space. 
    \item Comprehensive experiments demonstrate the state-of-the-art performance on multiple downstream tasks. 
\end{enumerate}

\section{Related Work}

\subsection{Rotation-Invariant Point Cloud Analysis}
Deep neural networks have become a cornerstone for learning point cloud representations, giving rise to numerous methods \cite{qi2017pointnet,qi2017pointnet++,wang2019dynamic,li2018pointcnn}. Despite their impressive success across a variety of downstream tasks, these methods exhibit a notable sensitivity to point cloud rotations. 
In response to the challenges posed by rotation perturbations, 
a number of rotation-robust methods have been proposed, based on techniques such as spherical convolutional neural networks and their variants \cite{esteves2018learning,You2021PRINSPRINOE}, regular icosahedral lattice \cite{Rao2019SphericalFC}, rotation conditions \cite{qiu2022spe}, tensor field networks \cite{poulenard2021functional} and adversarial training \cite{wang2022art}. 
These methods have marked encouraging improvements in handling rotated point clouds.  Nevertheless, 
these methods continue to face challenges in scenarios when confronted with data subjected to arbitrary rotations since they fail to obtain identical representations from point clouds in different poses.

To learn consistent features across a range of potential rotations, many methods \cite{chen2019clusternet,li2021rotation,gu2021learning} transform the input point cloud into rotation-invariant, handcrafted features, such as distances and angles. These features, derived from point coordinates and normal vectors, remain independent regarding the global orientation of point clouds. 
Drawing inspiration from 2D image representation learning, several methods \cite{chen2022devil,zhang2022riconv++} introduced rotation-invariant convolutions to extract point cloud features. 
However, by solely focusing on distances and angles, these methods may lose other vital information. To address this, some methods align point clouds to canonical poses, thereby learning rotation-invariant representations through the establishment of global reference frames \cite{li2021closer,Zhang2020GlobalCA,Lou2023CRINRP}. Unlike these existing approaches generally requiring complex feature engineering and network designs, 
our method eliminates the need for intricate feature or network designs with a simplified process, and allows for training without the necessity of annotating data.

\begin{figure*}[ht]
	\centering
			\includegraphics[width=\textwidth]{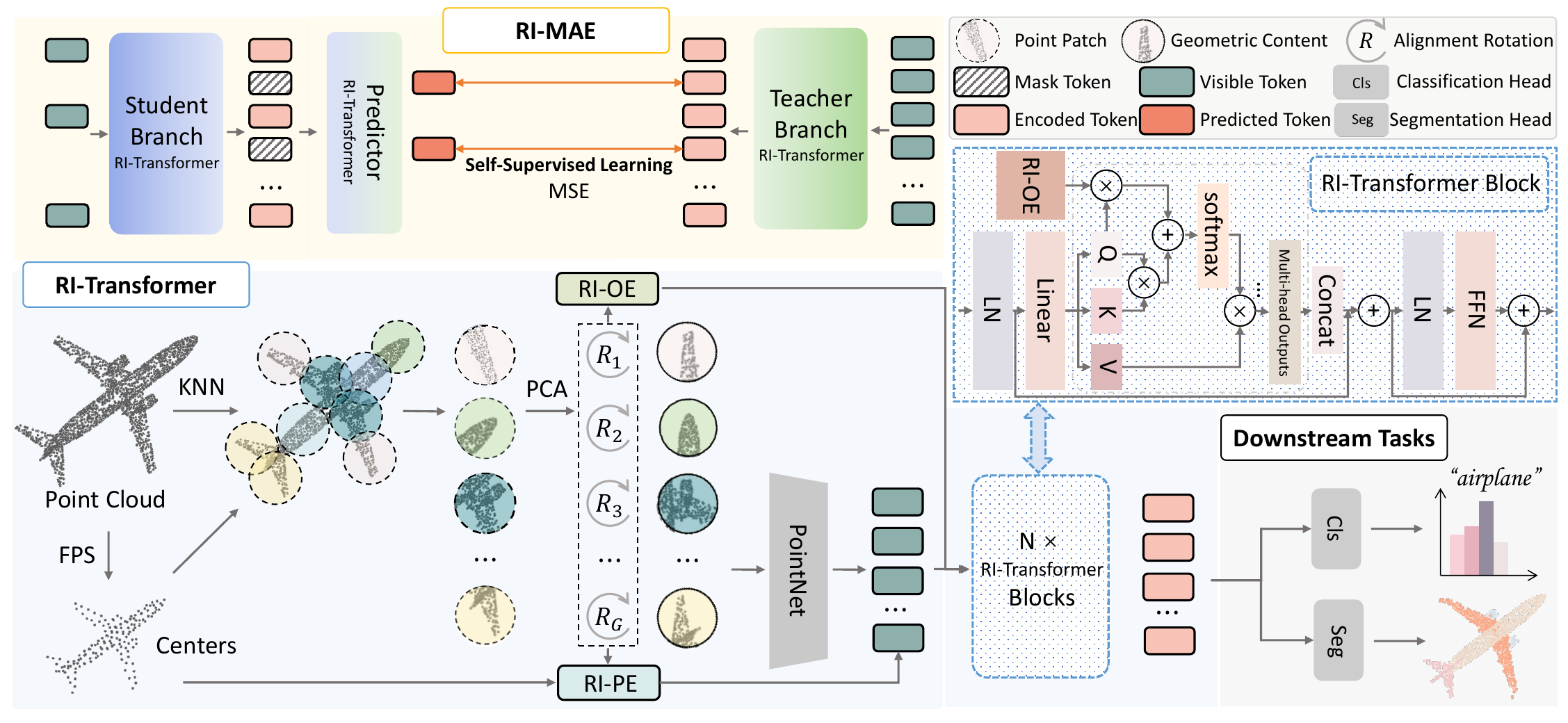}
	\caption{Overview of the proposed RI-MAE architecture. 
 The input point cloud is first divided into point patches via FPS and KNN, and PCA is utilized to align the patches and obtain rotation matrices relative to canonical poses. Then geometry content tokens, RI-OEs, and RI-PEs are formulated as RI-Transformer's inputs. Finally, task heads are employed for downstream tasks. A dual-branch student-teacher scheme is devised to conduct the self-supervised learning to pretrain the RI-Transformer. }
	\label{fig:supervised}
\end{figure*}

\subsection{Self-Supervised Learning for Point Clouds}
Self-supervised learning for point clouds has been explored extensively, leading to many promising approaches.  
These approaches train models with carefully designed pretext tasks instead of using labeled data.  
Common pretext tasks include point cloud reconstruction efforts such as input reconstruction \cite{yang2018foldingnet}, local-to-global reconstruction \cite{Liu2019L2GAU}, restoration of anomalous parts \cite{Chen2021ShapeSF}, and completion of occlusions \cite{Wang2020UnsupervisedPC}. Additionally, various contrastive learning approaches have been explored \cite{xie2020pointcontrast,rao2020global,sanghi2020info3d,afham2022crosspoint}, which learn point cloud representations by differentiating between pre-defined positive and negative samples.  
Other pretext tasks for self-supervised learning include orientation estimation \cite{Poursaeed2020SelfSupervisedLO}, mixing and disentangling techniques \cite{sun2022self}, and pose disentanglement strategies \cite{Tsai2022SelfSupervisedFL}.

Recently, inspired by the great achievement of BERT \cite{devlin2018bert} and MAE \cite{He2021MaskedAA}, many Transformer-based MPM methods for point cloud have been proposed and significantly outperform other methods in downstream tasks. Point-BERT \cite{yu2022point} introduced the BERT pretraining scheme to point clouds, which trains models with a task to predict discrete tokens generated by dVAE \cite{rolfe2016discrete}. Point-MAE \cite{pang2022masked} proposed a self-supervised pretext task to reconstruct masked input point patches based on the known tokens. Point-M2AE \cite{zhang2022point} employed a pyramid encoder and decoder design to conduct masked point cloud reconstruction in a hierarchical manner. ACT \cite{dong2023act}, I2P-MAE \cite{zhang2023learning}, ReCon \cite{qi2023recon} and ReCon++ \cite{qi2024shapellm} improved the 3D masked modeling by introducing pre-trained model from other modalities, such as images and natural languages. Point-GPT \cite{chen2024pointgpt} arranged point patches in an ordered sqeuence and extended the concept of genreative pre-training transformer (GPT) to point clouds.
Despite achieving remarkable results, these MPM methods are sensitive to rotation, constraining their scalability and generalizability.

\section{Method}

\subsection{Revisiting Masked Point Modeling}
We first revisit the masked point modeling (MPM) framework for point clouds. Existing methods, such as Point-MAE, segment the input point cloud \(\mathbf{X}\) into \(G\) point patches \(\mathbf{P}=\{\mathbf{p}_i\}_{i=1}^G\) using Farthest Point Sampling (FPS) and K-Nearest Neighborhood (KNN) algorithms. Subsequently, a masking ratio \(\alpha\) is applied, partitioning the point patches into a visible set \(\mathbf{P}_v=\{\mathbf{p}_i^v\}_{i=1}^{V}\) and a masked set \(\mathbf{P}_m=\{\mathbf{p}_i^m\}_{i=1}^{M}\), where \(V=(1-\alpha)\times G\) and \(M=\alpha \times G\) are the numbers of visible and masked patches, respectively. 
Following this, a mini-PointNet \cite{qi2017pointnet} is utilized to tokenize the visible point patches as point tokens \(\mathbf{T}_v=\{\mathbf{t}_i^v\}_{i=1}^{V}\) and the coordinates of patch centers are projected as position embeddings \(\{\mathbf{pos}_i\}_{i=1}^G\) via a learnable MLP (multi-layer perceptron). To this end, an encoding stage is introduced, during which only the visible tokens \(\mathbf{T}_v\) are fed into a transformer-based encoder:
\begin{align}
\label{eq:mpmencoder}
  \mathbf{T}_e = \mathrm{Encoder}(\mathbf{T}_v).
\end{align}
In the decoding stage, a full set of tokens consisting of encoded visible tokens \(\mathbf{T}_e\) and learnable mask tokens \(\mathbf{T}_m = \{\mathbf{t}^m_i\}_{i=1}^M\) are fed into a transformer-based decoder. 
Every masked token shares a learned masked embedding vector \(\mathbf{t}^m\), matching the dimension of \(\mathbf{t}_i^v\). The decoder's output for these masked tokens is denoted as \(\mathbf{H}^m = \{h^m_i\}_{i=1}^{M}\). Finally, a prediction head that consists of fully connected (FC) layers, formulates the point coordinates of the masked point patches \(\mathbf{P}_m\) based on \(\mathbf{H}^m\) as follows:
\begin{align}
\label{eq:mpmdecoder}
    \hat{\mathbf{P}}_m = \mathrm{FC}(\mathbf{H}^m) = \mathrm{FC}(\mathrm{Decoder}(\mathbf{T}_e, \mathbf{T}_m)),
\end{align} 
where \(\hat{\mathbf{P}}_m\) can be viewed as the estimation of  \(\mathbf{P}_m\). For simplicity, the position embeddings, which are integrated with the respective tokens in both the encoder and the decoder to indicate location context, are not depicted in Eq. (\ref{eq:mpmencoder}) - (\ref{eq:mpmdecoder}).

We will next explain our proposed method, which is illustrated in Fig. \ref{fig:supervised}. We first present a rotation-invariant transformer backbone, RI-Transformer for point clouds; and then a rotation-invariant MPM approach, RI-MAE, is introduced.

\subsection{Rotation-Invariant Transformer}
\label{sub:Rotation-Invariant Transformer}
We introduce a rotation-invariant transformer, namely RI-Transfomer, as an effective representation learning backbone for point clouds. The overall framework is shown in Fig. \ref{fig:supervised}. Our RI-Transfomer aims to decompose the point patches into 1) geometric content tokens and 2) orientation embeddings; and to provide location information with 3) rotation-invariant position embeddings. 

\paragraph{\textbf{Content-Orientation Disentanglement \& Content Tokenization}} 
\label{sub:Orientation Embedding}
This disentanglement mechanism allows our framework to learn and utilize the geometric content and orientation knowledge in a flexible manner. They can be further formulated in a transformer architecture, enhancing the pipeline's capabilities beyond what conventional MPM methods offer. 

For point cloud \(\mathbf{X}\), each of its point patches \(\mathbf{P}=\{\mathbf{p}_i\}_{i=1}^G\) can be viewed as having been rotated from a canonical pose with a specific rotation, i.e.,
\begin{align}
    \mathbf{P}=\{\mathbf{p}_i\}_{i=1}^G=\{\bar{\mathbf{p}}_i\mathbf{R}_i\}_{i=1}^G,
\end{align}
where \(\bar{\mathbf{p}}_i\) is the point patch \(\mathbf{p}_i\) in the canonical pose, and \(\mathbf{R}_i\) is the rotation that rotates \(\bar{\mathbf{p}}_i\) into \(\mathbf{p}_i\). 
As a result, a point patch can be decomposed into two components: the canonical content information \(\bar{\mathbf{p}}_i\), which is rotation-invariant, and the orientation information \(\mathbf{R}_i\). 

To obtain \(\bar{\mathbf{p}}_i\) and \(\mathbf{R}_i\), we conduct principal component analysis (PCA) to \(\mathbf{p}_i\), which is widely utilized for transforming point cloud into canonical pose \cite{li2021closer,Xie2023GeneralRI,yu2023rethinking,kim2020rotation,xiao2020endowing,yu2020deep}.
Next, we utilize a mini-PointNet and an MLP to encode \(\bar{\mathbf{p}}_i\) and \(\mathbf{R}_i\) into the rotation-invariant token \(\bar{\mathbf{t}}_i\) and the orientation embedding \(\mathbf{ori}_i\). respectively. Formally, we have: 
\begin{gather}
     \bar{\mathbf{p}}_i, \mathbf{R}_i = \mathrm{PCA}(\mathbf{p}_i), \\
    \bar{\mathbf{t}}_i=\mathrm{PointNet}(\bar{\mathbf{p}}_i), \\
    \mathbf{ori}_i=\mathrm{MLP}(\mathbf{R}_i).
\end{gather}

\paragraph{\textbf{Rotation-Invariant Orientation Embedding (RI-OE)}}
Importantly, the rotation \(\mathbf{R}_i\), which aligns \(\mathbf{p}_i\) to its canonical form \(\bar{\mathbf{p}}_i\), varies when the point set \(\mathbf{P}\) undergoes a rotation characterized by \(\mathbf{R}\). To elaborate, we have the following relationship:
\begin{align}
    \bar{\mathbf{p}}_i, \mathbf{R}_i^\prime = \mathrm{PCA}(\mathbf{p}_i\mathbf{R}), \mathbf{R}_i^\prime = \mathbf{R}_i\mathbf{R},
\end{align}
where $\mathbf{R}_i^\prime$ is the rotation that rotates the canonical \(\bar{\mathbf{p}}_i\) into the rotated point patch \(\mathbf{p}_i\mathbf{R}\). 
Hence, 
the orientation embedding \(\mathbf{ori}_i\) is affected as well, since the MLP itself does not inherently ensure rotation invariance.

Fully discarding orientation information and solely retaining geometric content 
intuitively maintain a rotation-invariant approach. However, this also results in the loss of relative orientation knowledge between point patches within a point cloud, and undermines the quality of point cloud representations. 
Fortunately, the relative orientation between two patches remains invariant to rotation $\mathbf{R}$ and can be conveniently expressed using our decomposed orientation component, as shown in Fig. \ref{fig:fig1}. 

In detail, 
the relative orientation between patches \(\mathbf{p}_i\) and \(\mathbf{p}_j\) can be formulated as \(\mathbf{R}_{ij}=\mathbf{R}_j\mathbf{R}_i^T\), where \(i, j \in \{1, 2, \cdots, G\}\). Upon applying the rotation \(\mathbf{R}\) to \(\mathbf{X}\) and \(\mathbf{P}\): \(\mathbf{X}^\prime = \mathbf{X}\mathbf{R}\) and 
\(\mathbf{P}^\prime = \mathbf{P}\mathbf{R}=\{\mathbf{p}_i\mathbf{R}\}_{i=1}^{G}=\{\mathbf{p}_i^\prime\}_{i=1}^{G}\) 
can be obtained. 
Subsequently, regarding \(\mathbf{p}^\prime_i\) and \(\mathbf{p}^\prime_j\), we have the relative orientation $\mathbf{R}^\prime_{ij}$ between  them: 
\begin{equation}
    \begin{split}
        \mathbf{R}^\prime_{ij} &= \mathbf{R}^\prime_{j}\mathbf{R}^{\prime T}_i
        = (\mathbf{R}_j\mathbf{R})(\mathbf{R}_i\mathbf{R})^T \\
        &= \mathbf{R}_j\mathbf{R}\mathbf{R}^T\mathbf{R}_i^T 
        = \mathbf{R}_j\mathbf{R}_i^T
        = \mathbf{R}_{ij}.
    \end{split}
\end{equation}
Therefore, we embed this relative patch orientation as the rotation-invariant information for characterizing the relations between patches. 
For this purpose, inspired by \cite{Wu2021RethinkingAI}, we devise a rotation-invariant orientation embedding (RI-OE) mechanism for self-attentions. Given the input token \(\bar{\mathbf{T}}=\{\bar{\mathbf{t}}_i\}_{i=1}^G\), its self-attention with RI-OE is formulated as:
\begin{align}
     \mathrm{Attention}(\mathbf{Q}, \mathbf{K}, \mathbf{V}) = \mathrm{softmax}(\frac{\mathbf{Q}\mathbf{K}^T+\mathbf{B}}{\sqrt{d_k}})\mathbf{V},
\end{align}
where \(\mathbf{Q}=\bar{\mathbf{T}}\mathbf{W}^Q\), \(\mathbf{K}=\bar{\mathbf{T}}\mathbf{W}^K\), \(\mathbf{V}=\bar{\mathbf{T}}\mathbf{W}^V\) are the query, key and value matrices; \(\mathbf{W}^Q, \mathbf{W}^K\) and \(\mathbf{W}^V\) are learnable matrices; and \(\mathbf{B} = \{b_{ij}\} \in \mathbb{R}^{G \times G}\) is RI-OE. We formulate \(b_{ij}\) by establishing bilinear relations that integrate the features of the $i^\text{th}$ patch token with the relative orientation information pertaining to the $j^\text{th}$ patch: 
 \begin{gather}
     b_{ij} = \bar{\mathbf{t}}_i\mathbf{W}^Q\mathbf{r}_{ij}^T, \\
     \mathbf{r}_{ij}=\mathrm{MLP}(\mathbf{R}_{ij}).
 \end{gather}
Note that for computational efficiency, we adopt the query weight matrix for the bilinear operator empirically. 
Instead of predicting the relative rotation with learnable module as done in existing method \cite{deng2023se}, our RI-OE directly employs rotation matrices to delineate the relative orientation, thereby avoiding intricate network designs.

\paragraph{\textbf{Rotation-Invariant Position Embedding (RI-PE)}}
Position embeddings offer critical location information for each point patch in MPM methods. Existing studies typically embed the coordinates of patch centers with MLPs. However, this practice is sensitive to rotations, as the coordinates of centres significantly alter with the rotation of point clouds.
To address this, we introduce the RI-PE mechanism. 

Specifically, for a point patch \(\mathbf{p}_i\) with center coordinate \(\mathbf{c}_i\), we utilize its rotation matrix \(\mathbf{R}_i\). 
The position embedding \(\mathbf{ripos}_i\) for \(\mathbf{p}_i\) can be formulated as:
\begin{align}
    \mathbf{ripos}_i = \mathrm{MLP}(\mathbf{c}_i\mathbf{R}_i^T),
\end{align}
where \(\mathbf{c}_i\mathbf{R}_i^T\) represents the location of the origin in the local reference frame of \(\mathbf{p}_i\). The rotation invariance of \(\mathbf{ripos}_i\) is discussed in the technical appendix.

To this end, given the proposed mechanisms including RI-OE and RI-PE, our novel transformer architecture, RI-Transformer, can be constructed with the property of rotation invariance. Compare with existing rotation-invatiant methods, our RI-Transformer effectively derives two embeddings rooted in content-orientation disentanglemen to preserve the intrinsic relative pose and local positional information. This helps avoid complex feature and network design as in existing methods, and maintain the capability to adopt any general transformer architectures seamlessly.

\subsection{Rotation-Invariant MAE}
\label{sub:Point Cloud rotation-invariant MAE}
Existing MPM methods utilize raw point clouds to guide the reconstruction of masked patches. However, this approach is not suitable for building a rotation-invariant MPM. Specifically, for a given point cloud in different directions, a rotation-invariant MPM should yield identical point cloud latent representations; however, the coordinates of the masked patches are different, necessitating that the decoder reconstructs distinct patches from the same latent representation. This conflict results in unstable loss, making the model difficult for optimization. We provide theoretical proof for this in the technical appendix.

Therefore, we introduce a rotation-invariant MAE architecture, RI-MAE, for rotation-invariant self-supervised learning. RI-MAE is with a dual-branch student-teacher architecture, as illustrated in Fig. \ref{fig:supervised}. It enables the reconstruction of masked patches within a learned rotation-invariant latent space instead of the rotation-sensitive input space, avoiding rotation interference. Overall, the two branches involves two RI-Transformers correspondingly and are connected with an additional RI-Transformer predictor. 
The teacher branch processes and encodes all point patches \(\mathbf{P}=\{\mathbf{p}_i\}_{i=1}^G\). In the student branch, 
only the visible set \(\mathbf{P}_v=\{\mathbf{p}_i^v\}_{i=1}^{V}\) is fed for encoding. 
Subsequently, the RI-Transformer predictor, with a single Transformer layser, is employed to predict the features of the masked point patches \(\bar{\mathbf{Z}}^t=\{\bar{\mathbf{z}}_i^t\}_{i=1}^{M}\). It uses the output embeddings of the visible point patches from the student branch, along with essential mask tokens, for the prediction. The supervision is based on the corresponding outputs from the teacher branch. The prediction is denoted as \(\bar{\mathbf{Z}}^s=\{\bar{\mathbf{z}}_i^s\}_{i=1}^{M}\). 
Given that the reconstruction process is carried out within the learned rotaion-invariant latent space, our dual-branch model guarantees the rotation invariance of the reconstruction. This attribute ensures the model's ability to pre-train effectively, even in scenarios where the orientation of the point cloud is unknown.

Note that the teacher and student branches are initialized randomly. The student updates its parameters through back-propagation, while the teacher operates as momentum-based, with its parameters being updated as the exponential moving average (EMA) of the student's parameters. 
Following the training of the dual-branch architecture, only the student encoder is utilized for downstream tasks. 

\paragraph{\textbf{Loss Function.}} Given the target masked point patch representations \(\bar{\mathbf{Z}}^t=\{\bar{\mathbf{z}}_i^t\}_{i=1}^{M}\) from the teacher branch and the predicted representations \(\bar{\mathbf{Z}}^s=\{\bar{\mathbf{z}}_i^s\}_{i=1}^{M}\) outputted from the predictor, a loss function is defined based on the mean squared error (MSE): 
\begin{equation}
    \begin{split}
    \mathcal{L} = \mathrm{MSE}(\bar{\mathbf{Z}}^t, \bar{\mathbf{Z}}^s) = \frac{1}{M}\sum_{i=1}^{M}||\bar{\mathbf{z}}_i^t-\bar{\mathbf{z}}_i^s||_2^2.
    \end{split}
\end{equation}

\section{Experiments}

\subsection{Implementation Details}
\label{sect:implementation details}
Following existing MPM methods, we utilized FPS and KNN to divide an input point cloud into \(G\) point patches with \(K=32\) points for each patch. For point cloud classification, we set \(G=64\), while for the segmentation tasks, \(G=256\). The RI-Transformer encoders in RI-MAE contain 12 transformer layers, while the predictor has only one transformer block. For each transformer block, we set the internal dimension to 384 and the number of heads to 6. 

For training details, we utilized an AdamW optimizer with a cosine learning rate decay, applying a decay factor 0.05, and incorporated a 10-epoch warm-up phase.
The learning rate was set at 0.0005 for both pretraining and classification, while for segmentation, it was adjusted to 0.0002. 
We pretrained RI-MAE on ShapeNet \cite{Chang2015ShapeNetAI}, which consists of more than 50,000 3D models, covering 55 categories. 
Following ClusterNet \cite{chen2019clusternet}, we conducted experiments in three scenarios to evaluate the rotation invariance: 1)  \textit{z/z} for training and testing with azimuthal rotations; 2) \textit{z/SO3} for training with azimuthal rotations and testing with arbitrary rotations; and 3) \textit{SO3/SO3}, entailing both training and testing under arbitrary rotations. All experiments were conducted utilizing two RTX 2080Ti GPUs, with the PyTorch framework version 1.7.

\subsection{Downstream Tasks}
\label{downstream tasks}

\setlength{\tabcolsep}{1pt}
\begin{table}[ht]
    \centering
    \begin{tabular}{lccc} 
    \toprule
        Methods &  \textit{z/z}&  \textit{SO3/SO3}&  \textit{z/SO3}\\
         \midrule
         \noalign{\smallskip}RaRI-Conv \cite{chen2022devil}&  83.3&  83.3&  83.3\\
 CRIN$\ddagger$ \cite{Lou2023CRINRP}& 84.7& 84.7& 84.7\\
 Yu et al. \cite{yu2023rethinking}& 86.6& 86.3& \underline{86.6}\\
 LocoTrans \cite{chen2024local}& 85.0& 84.5&85.0\\
 \midrule Transformer \cite{yu2022point}& 79.9& 58.7& 20.0\\
 Point-BERT \cite{yu2022point}& 87.4& 81.6& 21.5\\
 Point-MAE\(\dagger\) \cite{pang2022masked}& 88.3& 85.4& 23.8\\
 Point-M2AE\(\dagger\) \cite{zhang2022point}& 91.2& \underline{86.7}& 26.7\\
 ACT\(\dagger\) \cite{dong2023act}& \textbf{93.3}& 83.6& 22.7\\
 PointGPT-S\(\dagger\) \cite{chen2024pointgpt}& 91.7& 85.4&18.8\\
 \midrule
 RI-Transfomer& 88.3& 88.3& 88.3\\
 RI-MAE& 91.6& 91.6& 91.6\\ 
 RI-MAE\(\dagger\)& \underline{91.9}& \textbf{91.9}& \textbf{91.9}\\
 \bottomrule
    \end{tabular}
    \caption{Real-world classification accuracy (\%) on ScanObjectNN. \(\dagger\) indicates the method uses 2048 points as input. $\ddagger$ denotes additional normal vectors were fed.}
\label{tab:scanobjectnn_obj_bg}
\end{table}
\setlength{\tabcolsep}{1.4pt}

\setlength{\tabcolsep}{3pt}
\begin{table}[ht]
    \centering
    \begin{tabular}{lcc} 
    \toprule
        \multirow{2}{*}{Methods} &  \multicolumn{2}{c}{10-way}\\ 
           \cmidrule(lr){2-3}
         &  10-shot&  20-shot\\ 
         \hline\noalign{\smallskip}CrossPoint \cite{afham2022crosspoint}&  27.0 $\pm$ 2.9&  26.8 $\pm$ 2.7\\
 Transformer \cite{yu2022point}& 29.0 $\pm$ 5.6& 31.8 $\pm$ 3.3\\
 Point-BERT \cite{yu2022point}& \underline{36.5 $\pm$ 3.2}& 36.4 $\pm$ 2.2\\
 Point-MAE \cite{pang2022masked}& 36.4 $\pm$ 2.4& \underline{37.0 $\pm$ 1.9}\\
 Point-M2AE \cite{zhang2022point}& 35.9 $\pm$ 4.2& 36.7 $\pm$ 3.1\\
 ACT \cite{dong2023act}& 33.8 $\pm$ 6.7& 36.2 $\pm$ 4.7\\
 PointGPT-S \cite{chen2024pointgpt}& 32.6 $\pm$ 3.7 & 32.0 $\pm$ 2.7 \\
 \midrule
 RI-Transfomer& 78.9 $\pm$ 6.4& 86.8 $\pm$ 5.2\\
 RI-MAE& \textbf{90.2 $\pm$ 5.5}& \textbf{93.7 $\pm$ 3.5}\\ 
 \bottomrule
    \end{tabular}
    \caption{Few-shot object classification results (\%) on ModelNet40 in the \textit{z/SO3} scenario.}
\label{tab:fsl_10_way}
\end{table}
\setlength{\tabcolsep}{1.4pt}

\setlength{\tabcolsep}{1.5pt}

\begin{table}
\begin{center}
\begin{tabular}{lcc}
\toprule
Methods  & mIoU\(_I\)  & mIoU\(_C\)   \\
\midrule

 LGR-Net \cite{zhao2022rotation}& 82.8 &80.1\\

 RIFrame \cite{li2021rotation}& 82.5 &79.4\\
 Luo et al. \cite{Luo2022EquivariantPC}& -& 81.0\\

 Xie et al. \cite{Xie2023GeneralRI}& \underline{83.9} &\underline{81.5}\\

 RaRI-Conv \cite{chen2022devil}& 83.8 &-\\
 CRINet \cite{Lou2023CRINRP}& -& 80.5\\
 Yu et al. \cite{yu2023rethinking}& -& 80.3\\
 LocoTrans \cite{chen2024local}& 84.0 & 80.1 \\
\midrule
 Transformer \cite{yu2022point}& 29.1 &33.4\\
 Point-BERT \cite{yu2022point}& 29.2 &33.8\\
 Point-MAE \cite{pang2022masked}& 31.5 &33.8\\
 Point-M2AE \cite{zhang2022point}& 35.4 &38.7\\
 ACT \cite{dong2023act}& 35.7 &39.4\\
 PointGPT-S \cite{chen2024pointgpt}& 32.3&34.3\\
\midrule
 RI-Transfomer& 84.0 & 81.6\\
 RI-MAE& \textbf{84.3} &\textbf{82.1}\\
\bottomrule
\end{tabular}
\caption{Part segmentation on ShapeNetPart with mean IoU for all instances mIoU\(_I\) (\%) and mean IoU for all categories mIoU\(_C\) (\%), in the \textit{z/SO3} scenarios.}
\label{tab:shapenet_mean}
\end{center}
\end{table}
\setlength{\tabcolsep}{1.4pt}

\paragraph{\textbf{Real-Word Classification on ScanObjectNN.}} We evaluate the proposed method on a classification task with a real-word dataset - ScanOjectNN \cite{Uy2019RevisitingPC}, which contains 2,902 point clouds of 15 categories collected in real world. 
There are three variants of ScanOjectNN: \textit{OBJ\_BG}, \textit{OBJ\_ONLY}, and \textit{PB\_T50\_RS}. We evaluate our approach on the \textit{OBJ\_BG} variant by incorporating additional MLP classification heads, with 1,024 points sampled, and compare our method with state-of-the-art rotation-invariant and MPM approaches. It is worth emphasizing that, for the MPM methods, experiments in rotation scenarios were conducted utilizing the open-source code. The results are detailed in Table \ref{tab:scanobjectnn_obj_bg}.
Specifically, we trained an RI-Transfomer without utilizing the proposed dual-branch self-supervised learning strategy as a baseline.
The baseline RI-Transformer exhibits superior performance compared to the state-of-the-art rotation-invariant methods, notably outperforming the second-best approach \cite{yu2023rethinking} by a margin of +1.7\%. 
Regrading RI-MAE, with the dual-branch learning strategy, 
it delivers improvements of +2.6\% over the RI-Transfomer baseline. These results indicate the superiority of the proposed mechanisms. The results of the remaining two variants are included in the technical appendix.

\begin{figure}
	\centering
			\includegraphics[width=0.95\linewidth]{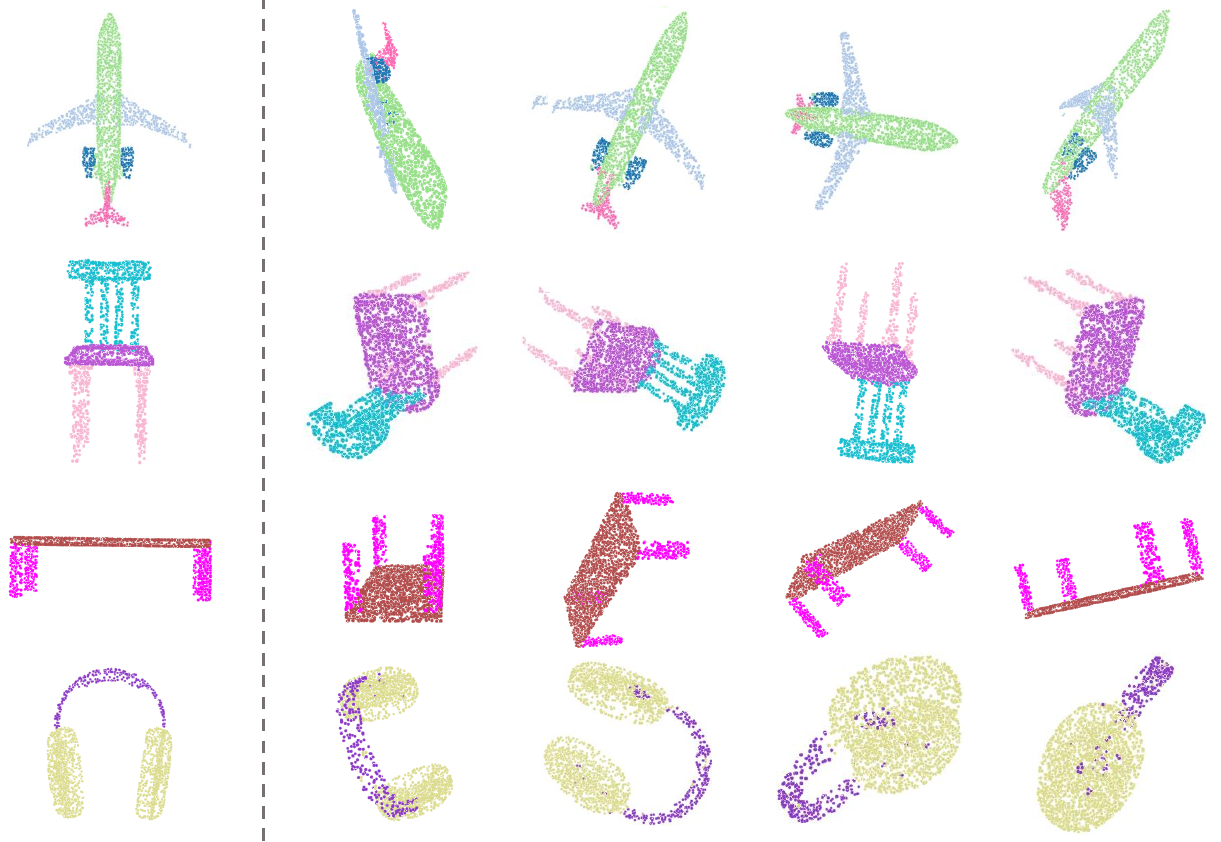}
	\caption{Visualization of part segmentation results on ShapeNet in the \textit{z/SO3} scenario. The leftmost column is the ground truth and the rest columns are the testing results of RI-MAE under arbitrary rotations.}
	\label{fig:seg_shapenet}
\end{figure}

\paragraph{\textbf{Few-shot Learning on ModelNet40.}} 
ModelNet40 is a synthetic shape dataset that contains 12,311 CAD models across 40 categories, with 1,024 points sampled from each model \cite{Wu20143DSA}. We conducted few-shot learning on ModelNet40, following practices from existing MPM methods. 
The performance is assessed across two settings: \textit{10-way 10-shot} and \textit{10-way 20-shot}, where \textit{\(K\)-way \(N\)-shot} indicates that \(N\) objects are sampled for training from each of \(K\) randomly selected classes, followed by testing on 20 unseen instances randomly chosen from each of these \(K\) classes. 
All experiments were conducted in the \textit{z/SO3} scenario to evaluate rotation robustness.
We report both the average accuracy and the standard deviation from 10 independent runs in Table \ref{tab:fsl_10_way}. 
The results show that our RI-MAE significantly outperforms existing methods, showing 
more than 50\% improvements. We also compare performance in other train/test scenarios and report the results of \textit{5-way 10-shot} and \textit{5-way 20-shot} settings in the technical appendix. 

\paragraph{\textbf{Part Segmentation on ShapeNet.}} 
The evaluation on part segmentation is with ShapeNet, which consists of 16,881 objects across 16 categories. In line with existing MPM methods, we sample 2,048 points from each object as input. 
A simple segmentation head is incorporated as in Point-MAE \cite{pang2022masked}.  
The segmentation results in the \textit{z/SO3} scenario are reported in Table \ref{tab:shapenet_mean}. It is evident that existing MPM methods are sensitive to rotations, resulting in significantly unsatisfactory accuracy levels. In contrast, our RI-MAE demonstrates robustness to rotations and achieves the best performance, with a mean intersection over union (mIoU) score of 84.3\% \(\mathrm{mIoU}_I\) and 82.1\% \(\mathrm{mIoU}_C\). 
In addition, the part segmentation results of RI-MAE in the \textit{z/SO3} scenario are visualized in Fig. \ref{fig:seg_shapenet}, indicating consistent predictions regardless of rotation. We also provide the results in the \textit{z/z} scenario in the technical appendix.

\setlength{\tabcolsep}{2pt}
\begin{table}
    \centering
\begin{tabular}{lccc}
    \toprule
         Methods&  \textit{z/z}&  \textit{SO3/SO3}& \textit{z/SO3}\\
            \midrule\noalign{\smallskip}
         GCAConv \cite{Zhang2020GlobalCA}&  -&  55.8& 55.7\\
         Xie et al. \cite{Xie2023GeneralRI}&  44.6&  44.6& 44.6\\
         LGR-Net \cite{zhao2022rotation}&  43.4&  43.4& 43.4\\
         Yu et al. \cite{yu2023rethinking}& 51.2 & 51.2 & 51.2\\
          LocoTrans \cite{chen2024local}& - & 56.0 & 54.2\\
            \midrule
         Transformer \cite{yu2022point}&  60.0&  57.1& 4.9\\
 Point-MAE \cite{pang2022masked}
& \underline{60.8}& 56.6&5.9\\
 ACT \cite{dong2023act}& \textbf{61.2}& 56.1&5.4\\
   \midrule
RI-Transfomer& 59.9& 59.9&59.9\\
        RI-MAE&  60.3&  \textbf{60.3}& \textbf{60.3}\\
         \bottomrule
    \end{tabular}
    \caption{Semantic segmentation accuracy with mIoU, \% on S3DIS under in three train/test scenarios on Area 5. }
\label{tab:s3dis}
\end{table}

\setlength{\tabcolsep}{3pt}
\begin{table}
    \centering
    \begin{tabular}{l|cc|cc|c}
    \hline
         Row&RI-OE & RI-PE & Dual-Branch & AE & Acc.\\
        \hline
        \#1&& &  \checkmark&  & 82.6\\
         \#2&\checkmark & &  \checkmark&  & 88.6\\
         \#3&& \checkmark&  \checkmark&  & 88.5\\
         \hline
         \#4&\checkmark &  \checkmark&  &  \checkmark& 86.7\\
         \hline
        \#5&\checkmark&  \checkmark&  \checkmark&  & \textbf{91.6}\\
         \hline
    \end{tabular}
    \caption{Ablation study on RI-OE, RI-PE and dual-branch.}
    \label{tab:ablation}
\end{table}

\paragraph{\textbf{Semantic Segmentation on S3DIS.}} 
We expanded our study to include challenging semantic segmentation on the large-scale 3D scenes dataset, S3DIS \cite{Armeni20163DSP}, which contains point clouds scanned from 271 rooms across six indoor areas, with 
all points annotated into 13 semantic categories. 
We sampled 4,096 points per room, utilizing only \(xyz\) coordinates as input, without incorporating \(rgb\) information. 
Areas 1 - 4 and 6  are for training and Area 5 is for testing. 
As detailed in Table \ref{tab:s3dis}, it is evident that our method outperforms other methods when tested with rotated data. It is noteworthy that our RI-Transformer significantly surpasses Yu et al. \cite{yu2023rethinking}, a heavily modified transformer, proving to be an effective backbone.

\subsection{Ablation Study}
\label{ablation stduy}
We conducted ablation studies to evaluate the effectiveness of the proposed mechanisms: 1) relative orientation embedding, 2) rotation-invariant position embedding and 3) the dual-branch architecture. 
We report the results on the \textit{OBJ\_BG} subset of the ScanObjectNN dataset, specifically under the \textit{z/SO3} scenario.

\paragraph{\textbf{Effectiveness of RI-OE and RI-PE.}} We first analyze the contribution of the proposed RI-OE and RI-PE. As shown in Table \ref{tab:ablation}, comparing the row \#1, \#2, \#3 and \#5, both our RI-OE and RI-PE deliver significant improvements and the best performance is achieved when the two mechanisms are employed. 
By incorporating the relative orientation and location information offered by RI-OE and RI-PE, our RI-MAE gains a deeper comprehension of the intricate structure within the input point cloud, facilitating the acquisition of a higher quality representation of the point cloud. 

\paragraph{\textbf{Effectiveness of the Dual-Branch.}} We also conducted an ablation study on the dual-branch architecture, where we replaced the dual-branch structure with an autoencoder (AE) architecture for pre-training. 
As evident in row \#4 and \#5 in Table \ref{tab:ablation}, our dual-branch architecture significantly surpasses the AE structure. As deliberated, when endeavoring AE to establish a rotation-invariant MPM framework, the uncertain reconstruction targets can lead to unstable loss and suboptimal solution. In contrast, our dual-branch architecture performs reconstructions within the learned rotation-invariant latent space, thus eliminating the influence of rotations and guaranteeing the efficacy of the pre-training.

\section{Conclusion}

In this paper, we present RI-MAE, a rotation-invariant masked point modeling approach, for self-supervised point cloud representation learning. 
RI-Transformer backbone is proposed to derive rotation-invariant latent space and a dual-branch design enables rotation-invariant masked point patches reconstruction.
Extensive experiments demonstrate the state-of-the-art performance of our method.

\bibliography{aaai25}

\appendix

\section{\textbf{Mathematical Proof}}
\paragraph{\textbf{Rotation Invariance of RI-PE.}} To confirm the rotation invariance of $\mathbf{ripos}_i$, RI-PE of the \(i^{th}\) point patch \(p_i\), 
consider when a point cloud \(\mathbf{X}\) undergoes rotation \(\mathbf{R}\), resulting in \(\mathbf{X}^\prime=\mathbf{X}\mathbf{R}\), the transformed point patches \(\mathbf{p}_i^\prime=\mathbf{p}_i\mathbf{R}\), alignment rotation matrix \(\mathbf{R}_i^\prime=\mathbf{R}_i\mathbf{R}\) and the center \(\mathbf{c}_i^\prime=\mathbf{c}\mathbf{R}\). Subsequently, we have:
\begin{equation}
\begin{split}
    \mathbf{c}_i^\prime \mathbf{R}_i^{\prime T} &= \mathbf{c}_i\mathbf{R}(\mathbf{R}_i\mathbf{R})^T\\
    &= \mathbf{c}_i\mathbf{R}\mathbf{R}^T\mathbf{R}_i^T\\ &= \mathbf{c}_i\mathbf{R}_i^T, 
\end{split}
\end{equation}
thus the RI-PE remains consistent after rotation:
\begin{equation}
\begin{split}
    \mathbf{ripos}_i^\prime &= \mathrm{MLP}(\mathbf{c}_i^\prime \mathbf{R}_i^{\prime T}) \\ &= \mathrm{MLP}(\mathbf{c}_i\mathbf{R}_i^T)\\
    &= \mathbf{ripos}_i.
\end{split}
\end{equation}

\paragraph{\textbf{Inadequacy and Unstableness of AE Structure.}}In Sec. I and III of the main paper, we have elaborated on the inadequacy of the autoencoder (AE) structure of existing MPM methods for constructing rotation-invariant MPMs, primarily due to their propensity to reconstruct unstable target. Here we present a theoretical justification to substantiate our claim. 

We denote \(\mathbf{T}_e\), \(\mathbf{T}_m\), \(\mathbf{P}_m\), \(\hat{\mathbf{P}}_m\) and \(l_{rec}\) as the encoder output, learnable mask tokens, masked point patches, the reconstructed point patches and the resulting loss, and the ``\(\prime\)" notation signifies the corresponding representation after the rotation of the input. Before rotation, \(\hat{\mathbf{P}}_m\) and \(l_{rec}\) can be obtained according to the following equations: 
\begin{gather}
     \hat{\mathbf{P}}_m=\mathrm{FC}(\mathrm{Decoder}(\mathbf{T}_e, \mathbf{T}_m)), \\
    l_{rec}=\mathrm{loss}(\hat{\mathbf{P}}_m,\mathbf{P}_m).
\end{gather}
After rotation, we have:
\begin{gather}
     \hat{\mathbf{P}}_m^\prime=\mathrm{FC}(\mathrm{Decoder}(\mathbf{T}_e^\prime, \mathbf{T}_m)), \\
    l_{rec}^\prime=\mathrm{loss}(\hat{\mathbf{P}}_m^\prime,\mathbf{P}_m^\prime).
\end{gather}
Since the learned representation is rotation invariant, i.e. \(\mathbf{T}_e^\prime=\mathbf{T}_e\), then we derive:
\begin{gather}
    \hat{\mathbf{P}}_m^\prime=\mathrm{FC}(\mathrm{Decoder}(\mathbf{T}_e, \mathbf{T}_m))=\hat{\mathbf{P}}_m, \\
    l_{rec}^\prime=\mathrm{loss}(\hat{\mathbf{P}}_m,\mathbf{P}_m^\prime).
\end{gather}
Since \(\mathbf{P}_m\neq\mathbf{P}_m^\prime\), the two losses are different: \(l_{rec}\neq l_{rec}^\prime\).

Hence, when applied to the same point cloud oriented in varying directions, the conventional AE architecture tends to yield disparate losses, ultimately resulting in significant instability during the training process.

\section{Additional Experiments}

\setlength{\tabcolsep}{1pt}

\begin{table}
\begin{center}
\resizebox{\linewidth}{!}{
\begin{tabular}{lccll}
\toprule
\multirow{2}{*}{Methods} & \multicolumn{2}{c}{z/z} & \multicolumn{2}{c}{z/SO3} \\
\cmidrule(lr){2-3} \cmidrule(lr){4-5}
& mIoU\(_I\)  & mIoU\(_C\)    &  mIoU\(_I\)  
&mIoU\(_C\)    
\\
\midrule

 LGR-Net \cite{zhao2022rotation}& -&-& 82.8 
&80.1 
\\

 RIFrame \cite{li2021rotation}& -&-& 82.5 
&79.4 
\\
 Luo et al. \cite{Luo2022EquivariantPC}& -& -& -
&81.0 
\\

 Xie et al. \cite{Xie2023GeneralRI}& 83.9&81.5& \underline{83.9} 
&\underline{81.5} 
\\

 RaRI-Conv \cite{chen2022devil}& -&-& 83.8 
&- 
\\
 CRINet \cite{Lou2023CRINRP}& -& 80.5 & -
&80.5 
\\
 Yu et al. \cite{yu2023rethinking}& -& -& -
&80.3 
\\
 LocoTrans \cite{chen2024local}& -& -& 84.0 
&80.1  
\\
\midrule
 Transformer \cite{yu2022point}& 85.1&83.4& 29.1 
&33.4 
\\
 Point-BERT \cite{yu2022point}& 85.6&84.1& 29.2 
&33.8 
\\
 Point-MAE \cite{pang2022masked}& 86.1&84.2& 31.5 
&33.8 
\\
 Point-M2AE \cite{zhang2022point}& \textbf{86.5}&\textbf{84.7}& 35.4 
&38.7 
\\
 ACT \cite{dong2023act}& 86.1&\textbf{84.7}& 35.7 
&39.4 
\\
 PointGPT-S \cite{chen2024pointgpt}& \underline{86.2}&84.1& 32.3
&34.3 
\\
\midrule
 RI-Transfomer& 84.0 & 81.6 & 84.0 
&81.6 
\\
 RI-MAE& 84.3&82.1& \textbf{84.3} &\textbf{82.1} \\
\bottomrule
\end{tabular}
}
\caption{Part segmentation on ShapeNetPart with mean IoU for all instances mIoU\(_I\) (\%) and mean IoU for all categories mIoU\(_C\) (\%), in the \textit{z/z} and \textit{z/SO3} scenarios.}
\label{tab:shapenet_meanv2}
\end{center}
\end{table}
\setlength{\tabcolsep}{1.4pt}

\setlength{\tabcolsep}{3pt}
\begin{table*}[ht]
    \centering
    \begin{tabular}{lccccccccc} 
    \toprule
        \multirow{2}{*}{Methods} &  \multicolumn{3}{c}{\textit{OBJ\_BG}}&  \multicolumn{3}{c}{\textit{OBJ\_ONLY}}&  \multicolumn{3}{c}{\textit{PB\_T50\_RS}}\\ 
         \cmidrule(lr){2-4}  \cmidrule(lr){5-7}  \cmidrule(lr){8-10}
         &  \textit{z/z}&  \textit{SO3/SO3}&  \textit{z/SO3}&  \textit{z/z}&  \textit{SO3/SO3}&  \textit{z/SO3}&  \textit{z/z}&  \textit{SO3/SO3}& \textit{z/SO3}\\ 
         \hline\noalign{\smallskip}RaRI-Conv \cite{chen2022devil}&  83.3&  83.3&  83.3&  -&  -&  -&  -&  -& -\\
 CRIN$\ddagger$ \cite{Lou2023CRINRP}& 84.7& 84.7& 84.7& -& -
& -
& -
& -
&-
\\
 Yu et al. \cite{yu2023rethinking}& 86.6& 86.3& 
 \underline{86.6}& -& -& -& -& -&-\\
 RI-Conv++ \cite{zhang2022riconv++}& 85.6& 85.6& 85.6& 86.2& \underline{86.2}& \underline{86.2}& 80.3& 80.3&\underline{80.3}\\
  LocoTrans \cite{chen2024local}& 85.0 & 84.5 & 85.0 & -&-&-&-&-&-\\
 \midrule Transformer \cite{yu2022point}& 79.9& 58.7& 20.0& 80.6& 62.0& 18.1& 77.2& 75.7&22.2\\
 Point-BERT \cite{yu2022point}& 87.4& 81.6& 21.5& 88.1& 84.9& 20.5& 83.1& 79.7&19.8\\
 Point-MAE\(\dagger\) \cite{pang2022masked}& 88.3& 85.4& 23.8& \underline{90.0}& 85.9& 17.0& 84.5& \underline{81.7}&19.3\\
 Point-M2AE\(\dagger\) \cite{zhang2022point}& 91.2& \underline{86.7}& 26.7& 88.8& 84.0& 24.8& \underline{86.4}& 80.8&24.6\\
 ACT\(\dagger\) \cite{dong2023act}& \textbf{93.3}& 83.6& 22.7& \textbf{91.0}& 83.0& 23.8& \textbf{88.2}& 80.6&21.1\\
 PointGPT\(\dagger\) \cite{chen2024pointgpt}& 91.7& 85.4& 18.8& 91.0& 85.4& 22.9& 86.9& 81.1&18.4\\
 \midrule
 RI-Transfomer& 88.3& 88.3
& 88.3
& 86.2& 86.2& 86.2& 82.4& 82.4
&82.4\\
 RI-MAE& 91.6& 91.6& 91.6& 88.8& 88.8& 88.8& 85.1& 85.1&85.1\\ 
 RI-MAE\(\dagger\)& \underline{91.9}& \textbf{91.9}& \textbf{91.9}& 89.2& \textbf{89.2}& \textbf{89.2}& 85.8& \textbf{85.8}&\textbf{85.8}\\
 \bottomrule
    \end{tabular}
    \caption{Real-world point cloud classification accuracy (\%) on ScanObjectNN in three scenarios. We test on the \textit{OBJ\_BG}, \textit{OBJ\_ONLY}, and \textit{PB\_T50\_RS} variants. \(\dagger\) indicates the method uses 2048 points as input. $\ddagger$ denotes additional normal vectors fed to the model.}
\label{tab:scanobjectnn}
\end{table*}
\setlength{\tabcolsep}{1.4pt}

\paragraph{\textbf{Real-Word Classification on ScanObjectNN.}} There are three variants of ScanObjectNN \textit{OBJ\_BG}, \textit{OBJ\_ONLT}, and \textit{PB\_T50\_RS}. Besides \textit{OBJ\_BG}, we also evaluate the proposed approach on the remaining two variants and compare with state-of-the-art methods in Table \ref{tab:scanobjectnn}. It is evident that the proposed method maintains rotation invariance across all three variances. Furthermore, in comparison with existing state-of-the-art MPM methods, our proposed approach demonstrates comparable performance in the \textit{z/z} scenario, while significantly outperforming these methods in the rotational testing scenario, underscoring its effectiveness and robustness.

\setlength{\tabcolsep}{1.5pt}
\begin{table*}[ht]
    \centering
    \begin{tabular}{lccccllll} 
    \toprule
        \multirow{3}{*}{Methods} & \multicolumn{4}{c}{z/z} & \multicolumn{4}{c}{z/SO3} \\
        \cmidrule(lr){2-5} \cmidrule(lr){6-9}
        &  \multicolumn{2}{c}{5-way}&  \multicolumn{2}{c}{10-way
} & \multicolumn{2}{c}{5-way} & \multicolumn{2}{c}{10-way}\\ 
         \cmidrule(lr){2-3}  \cmidrule(lr){4-5} \cmidrule(lr){6-7} \cmidrule(lr){8-9} 
         &  10-shot&  20-shot&  10-shot&  20-shot
 & 10-shot&  20-shot&  10-shot&  20-shot\\ 
         \hline\noalign{\smallskip}CrossPoint \cite{afham2022crosspoint}&  92.5 $\pm$ 3.0&  94.9 $\pm$ 2.1&  83.6 $\pm$ 5.3&  87.9 $\pm$ 4.2& 36.2 $\pm$ 9.1& 38.8 $\pm$ 10.7& 27.0 $\pm$ 2.9&26.8 $\pm$ 2.7
\\
 OcCo \cite{Wang2020UnsupervisedPC}& 90.6 $\pm$ 2.8& 92.5 $\pm$ 1.9& 82.9 $\pm$ 1.3& 86.5 $\pm$ 2.2&  40.2 $\pm$ 3.6& 41.7 $\pm$ 3.0& 23.6 $\pm$ 3.4&26.1 $\pm$ 2.2
\\
 Transformer \cite{yu2022point}& 87.8 $\pm$ 5.2& 93.3 $\pm$ 4.3& 84.6 $\pm$ 5.5& 89.4 $\pm$ 6.3&  41.5 $\pm$ 7.2& 46.4 $\pm$ 5.5& 29.0 $\pm$ 5.6&31.8 $\pm$ 3.3
\\
 Point-BERT \cite{yu2022point}& 94.6 $\pm$ 3.1& 96.3 $\pm$ 2.7& 91.0 $\pm$ 5.4& 92.7 $\pm$ 5.1&  48.0 $\pm$ 4.6& 48.8 $\pm$ 8.8& \underline{36.5 $\pm$ 3.2}&36.4 $\pm$ 2.2
\\
 Point-MAE \cite{pang2022masked}& \underline{96.3 $\pm$ 2.5}& \underline{97.8 $\pm$ 1.8}& \textbf{92.6 $\pm$ 4.1}& \textbf{95.0 $\pm$ 3.0}&  45.7 $\pm$ 4.1& 46.4 $\pm$ 7.3& 36.4 $\pm$ 2.4&\underline{37.0 $\pm$ 1.9}
\\
 Point-M2AE \cite{zhang2022point}& \textbf{96.8 $\pm$ 1.8}& \textbf{98.3 $\pm$ 1.4}& \underline{92.3 $\pm$ 4.5}& \textbf{95.0 $\pm$ 3.0}&  \underline{55.1 $\pm$ 8.3}& \underline{54.3 $\pm$ 6.2}& 35.9 $\pm$ 4.2&36.7 $\pm$ 3.1
\\
 ACT \cite{dong2023act}& 95.9 $\pm$ 2.2& 97.7 $\pm$ 1.8& 92.4 $\pm$ 5.0& 94.7 $\pm$ 3.9&  94.7 $\pm$ 3.9& 49.1 $\pm$ 4.6& 33.8 $\pm$ 6.7&36.2 $\pm$ 4.7
\\
 \midrule
 RI-Transfomer& 88.0 $\pm$ 6.1& 92.1 $\pm$ 3.2& 78.9 $\pm$ 6.4& 86.8 $\pm$ 5.2
 &  88.0 $\pm$ 6.1& 92.1 $\pm$ 3.2& 78.9 $\pm$ 6.4&86.8 $\pm$ 5.2
\\
 RI-MAE& 95.3 $\pm$ 3.5& 97.7 $\pm$ 1.5& 90.2 $\pm$ 5.5& 93.7 $\pm$ 3.5 &  \textbf{95.3 $\pm$ 3.5}& \textbf{97.7 $\pm$ 1.5}& \textbf{90.2 $\pm$ 5.5}&\textbf{93.7 $\pm$ 3.5} \\ 
 \bottomrule
    \end{tabular}
    \caption{Few-shot object classification results (\%) on ModelNet40 in the \textit{z/z} and \textit{z/SO3} scenarios. We conduct 10 independent experiments and report the mean accuracy(\%) and  the standard deviation.}
\label{tab:fsl}
\end{table*}
\setlength{\tabcolsep}{1.4pt}

\paragraph{\textbf{Few-shot Learning on ModelNet40.}} Besides \textit{10-way 10-shot} and \textit{10-way 20-shot}, we also conduct experiments with \textit{5-way 10-shot} and \textit{5-way 20-shot} settings in the \textit{z/z} and \textit{z/SO3} scenarios, and compare with existing MPM methods. As evidenced in Table \ref{tab:fsl}, our approach demonstrates robustness to rotation and significantly outperforms other MPM methods in the \textit{z/SO3} scenario, with more than 40\% improvements.

\paragraph{\textbf{3D Part Segmentation.}} Apart from the \textit{z/SO3} scenario, we have also conducted experiments in the \textit{z/z} scenario, and reported the mIoU across all classes and all instances in Table \ref{tab:shapenet_meanv2}. The consistent results obtained in both scenarios demonstrate the rotation invariance of our method.

\end{document}